\documentclass[a4paper,fleqn]{cas-dc}

\usepackage{float}
\usepackage[numbers]{natbib}
\usepackage{caption}
\usepackage{url}
\def\tsc#1{\csdef{#1}{\textsc{\lowercase{#1}}\xspace}}
\tsc{WGM}
\tsc{QE}
\usepackage{caption}
\begin{document}
\let\WriteBookmarks\relax
\def\floatpagepagefraction{1}
\def\textpagefraction{.001}

\shorttitle{}    

\shortauthors{}  

\title [mode = title]{Tree Learning: A Multi-Skill Continual Learning Framework for Humanoid Robots}  



% ========== 标题注释（通常留空或填基金信息）==========

% ========== 第一作者 ==========
\author[1]{Yifei Yan}  % 姓名，ORCID可选

% \cormark[1]           % 标记为通讯作者（会显示*号）
% \fnmark[1]            % 作者脚注标记（用于单位注释等，通常留空）

\ead{yyf20041019@shu.edu.cn}        % 邮箱（必填）

\credit{Methodology, Software, Writing - original draft, Visualization}  % 作者贡献

\affiliation[1]{
    organization={School of Future Technology, Shanghai University},
    addressline={No. 99 Shangda Road, Baoshan District}, 
    city={Shanghai},
    postcode={200444}, 
    state={Shanghai},
    country={China}
}

% ========== 第二作者（如果有导师/合作者）==========
\author[1]{Linqi Ye}

% \fnmark[1]            % 如果第二作者也是通讯作者，加 \cormark[2]，否则不加
\cormark[1]
\ead{yelinqi@shu.edu.cn}

\credit{Conceptualization, Supervision, Project administration, Funding acquisition, Writing - review \& editing}

% ========== 通讯作者说明文字 ==========
\cortext[1]{Corresponding author.}  % 这个会自动显示在页脚

% ========== 作者脚注文字（如果有特殊说明）==========
% \fntext[1]{}  % 通常留空，如果有说明就填（如"These authors contributed equally."）

% ========== 无编号标题注释 ==========
\nonumnote{Project page: \url{https://yyf-prog.github.io/Tree-learning/}}  % 如果需要就取消注释


\begin{abstract}
As reinforcement learning for humanoid robots evolves from single-task to multi-skill paradigms, efficiently expanding new skills while avoiding catastrophic forgetting has become a key challenge in embodied intelligence. Existing approaches either rely on complex topology adjustments in Mixture-of-Experts (MoE) models or require training extremely large-scale models, making lightweight deployment difficult. To address this, we propose Tree Learning, a multi-skill continual learning framework for humanoid robots. The framework adopts a root-branch hierarchical parameter inheritance mechanism, providing motion priors for branch skills through parameter reuse to fundamentally prevent catastrophic forgetting. A multi-modal feedforward adaptation mechanism combining phase modulation and interpolation is designed to support both periodic and aperiodic motions. A task-level reward shaping strategy is also proposed to accelerate skill convergence. Unity-based simulation experiments show that, in contrast to simultaneous multi-task training, Tree Learning achieves higher rewards across various representative locomotion skills while maintaining a 100\%\ skill retention rate, enabling seamless multi-skill switching and real-time interactive control. We further validate the performance and generalization capability of Tree Learning on two distinct Unity-simulated tasks: a Super Mario-inspired interactive scenario and autonomous navigation in a classical Chinese garden environment. 
\end{abstract}


%\nocite{*}

% Keywords

\maketitle

% Main text
\section{Introduction}\label{intro}

The rapid development of embodied intelligence is driving autonomous robot technology from controlled laboratory environments toward general-purpose real-world scenarios \citep{Wuetal2023,Zhuetal2024,Leeetal2020}. As a typical representative of embodied intelligence, humanoid robots possess significant advantages in navigating unstructured environments—such as roads, staircases, and debris—due to their human-like gait and body structure \citep{Haetal2025,LiuTian2011}. In recent years, reinforcement learning (RL)-based motion control methods \citep{XieChen2025,GouGuo2025,LiuLi2022,ZhangXia2025} have been evolving from single-task adaptation toward multi-skill incremental expansion, posing significant challenges in terms of generalization capability, continual learning efficiency, and system scalability.

In the domain of humanoid robots acquiring diverse skills, several recent advances have been achieved. BeyondMimic \citep{Liaoetal2025} realized flexible cross-task trajectory synthesis through a guided diffusion model. The KungfuBot series \citep{Xieetal2025,Hanetal2025} leveraged orthogonal mixture-of-experts (OMoE) models and bilevel optimization algorithms, enabling robots to imitate highly dynamic behaviors such as martial arts and dance. Any2Track \citep{ZhangGuo2025} proposed a two-stage framework that substantially enhanced motion tracking robustness under multi-source disturbances.

However, three major technical gaps remain with respect to the core demand for continual incremental expansion. First, existing MoE-based approaches \citep{ObandoCeron2024,Yangetal2020,Huangetal2025,Chengetal2023} often require network topology adjustments when introducing new skills, while approaches represented by SONIC \citep{Luoetal2025} tend to train extremely large-scale models, leading to computational costs that grow rapidly with the number of skills and making edge-side lightweight deployment difficult. Second, motion imitation-based methods \citep{Pengetal2018,Lietal2024,ZhangTang2025} typically require full policy retraining when learning new motions, which readily leads to degradation of existing fundamental capabilities—the well-known catastrophic forgetting problem. Third, existing robust control strategies \citep{ZhangXiao2024,Bellegarda2024,ChenCui2024} are largely limited to parameter adaptation for basic gaits and cannot achieve cross-modal motion generalization at low training cost.

\vspace{1em}
\begin{center}
\includegraphics[width=1\linewidth]{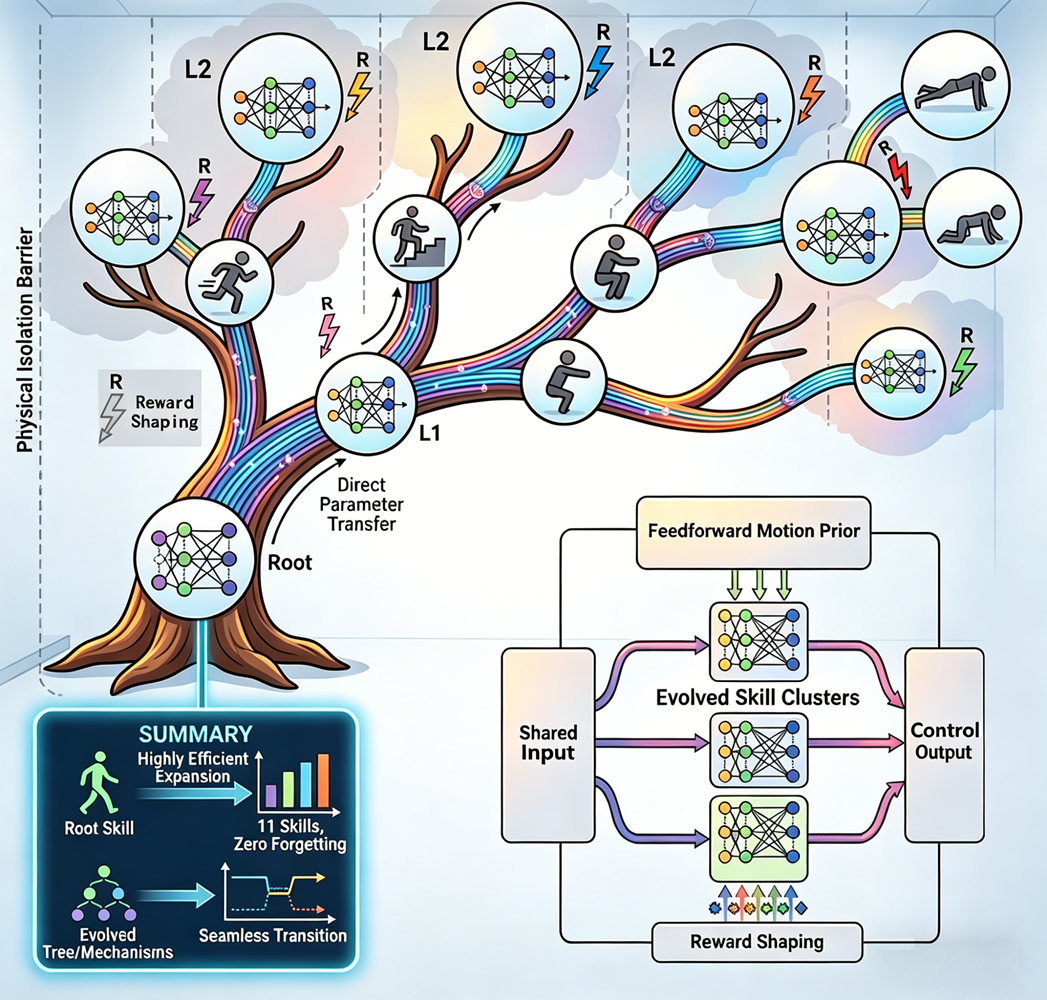}
\captionof{figure}{Tree Learning diagram.}
\label{fig0}
\end{center}
\vspace{1em}

Inspired by the mechanisms of prior inheritance and specialized evolution observed in biological evolution, this paper proposes Tree Learning (Figure \ref{fig0})—a multi-skill continual learning framework based on hierarchical inheritance. The framework defines a pre-trained flat-ground basic gait as the root skill, while other skills inherit root skill parameters in a tree-structured manner layer by layer, referred to as branch skills. Through parameter inheritance and reward shaping, branch skills achieve rapid evolution. The main contributions of this paper are as follows:

(1) A tree-structured parameter inheritance architecture for continual learning is proposed. Through physically isolated sub-network clusters, catastrophic forgetting is fundamentally eliminated, enabling lossless incremental expansion of multiple skills.

(2) A low-cost skill iteration pipeline is designed, supporting incremental fine-tuning of specific branches and significantly reducing computational and time costs.

(3) Simulation validation is conducted on the Unitree G1 humanoid robot, achieving efficient expansion and seamless switching of 11 types of highly dynamic skills with real-time interactive control.

\section{System Overview}\label{overview}

As shown in Figure \ref{fig2}, the Tree Learning architecture defines flat-ground walking as the root skill and extends branch skills layer by layer based on a hierarchical inheritance mechanism. Continuing training from the root skill yields first-level branch skills: one-leg standing, stair climbing, lying prone, running, and squatting. From one-legged standing, a second-level branch skill—ball kicking—is derived through further training. From squatting, a second-level branch skill—two-footed jumping—is obtained. From the lying prone action, second-level branch skills including push-ups, crawling, and standing up are developed.

\vspace{1em}
\begin{center}
\includegraphics[width=1\linewidth]{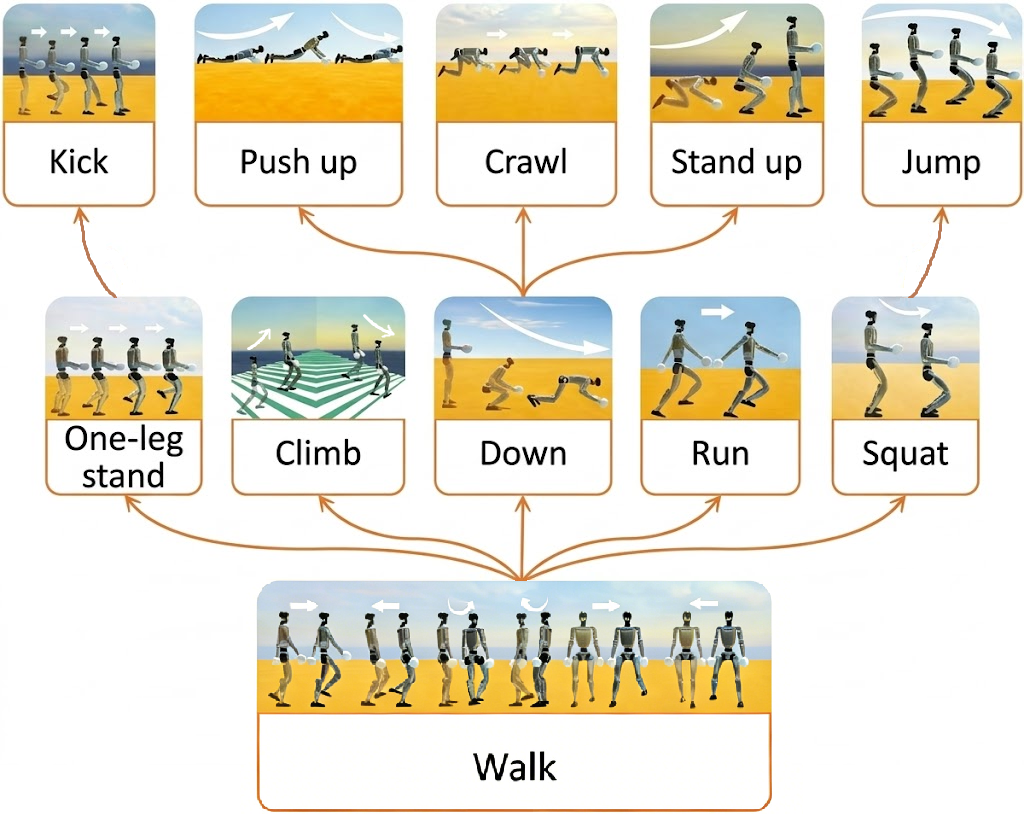}
\captionof{figure}{Tree Learning for Unitree G1.}
\label{fig2}
\end{center}
\vspace{1em}

The Tree Learning framework enforces consistency of the global state space and action representation by design. Since the root skill and all branch sub-networks share identical sensor input interfaces and joint control output dimensions, and network switching is only performed between skills with overlapping action sequences and state spaces, the system maintains high consistency in motion continuity during dynamic model loading and skill switching. This consistency fundamentally avoids action discontinuities at the moment of neural network switching, thereby effectively preventing abrupt posture changes of the robot and enabling seamless, smooth transitions between heterogeneous locomotion modes.

In complex terrain or task scenarios requiring fine-grained control, the Tree Learning framework supports user intervention in skill switching and action adjustment through a lightweight input interface. For example, users can select target skill models via numeric keys 0/1/2/3 (e.g., 0 for flat-ground walking, 1 for running) and dynamically adjust action commands via W/S/A/D/Q/E letter keys (e.g., W/S for forward/backward, A/D for left/right turning, Q/E for lateral movement). Upon receiving user commands, the system combines state assessment to load the corresponding model and output action commands, forming a closed-loop interaction of user input–model execution–state feedback. This human-robot collaborative mode significantly enhances system adaptability in dynamic scenarios.
\vspace{1em}
\begin{center}
\includegraphics[width=1\linewidth]{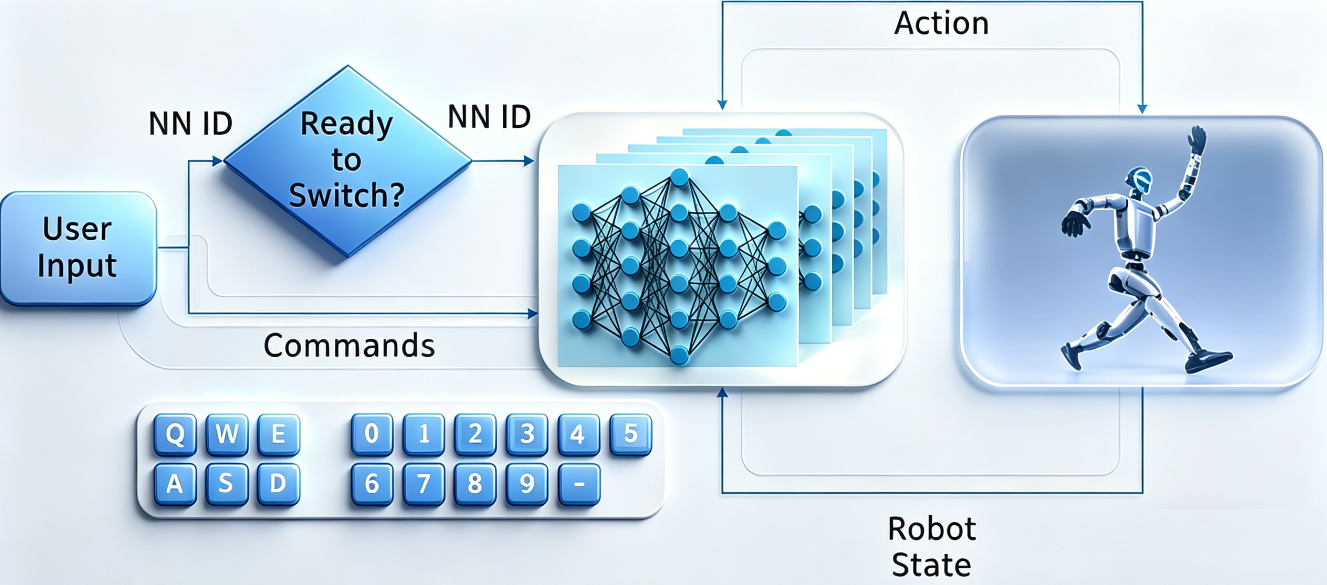}
\captionof{figure}{Interactive multi-skill control.}
\label{fig3}
\end{center}
\vspace{1em}

\section{Methodology}\label{method}

\subsection{Feedforward Action Design}
For each skill, motion prior is used as a feedforward action to achieve a specific motion style. Different actions are realized through simple feedforward signals. For periodic and aperiodic actions, phase-based and interpolation-based methods are employed, respectively.

\subsubsection{Phase modulation method}

For periodic locomotion skills such as walking, running, stair climbing, and jumping, a cosine phase modulation method is used to design the feedforward action trajectory. The core mechanism employs gait phase activation factors $u_1$ and $u_2$ to schedule alternating motion of the left and right legs of the humanoid robot, providing a baseline action reference for reinforcement learning and reducing exploration cost.

The factors $u_1$ and $u_2$ represent the gait phase activation factors for the robot's left and right legs, respectively. Their computation relies on the gait cycle counter $tp$ and the gait cycle period $T_1$.

The core formulas are given as follows:
\begin{equation}
\left\{
\begin{aligned}
& tp_0 =
\begin{cases}
tp, & 0 < tp \leq T_1 \\
tp - T_1, & T_1 < tp \leq 2T_1
\end{cases} \\[1.5em]
& u_1 =
\begin{cases}
\dfrac{1}{2}\left({1 - \cos\left(\dfrac{2\pi tp_0}{T_1}\right)}\right), & 0 < tp \leq T_1 \\[1.5em]
0, & T_1 < tp \leq 2T_1
\end{cases} \\[1.5em]
& u_2 =
\begin{cases}
0, & 0 < tp \leq T_1 \\[1.5em]
\dfrac{1}{2}\left({1 - \cos\left(\dfrac{2\pi tp_0}{T_1}\right)}\right), & T_1 < tp \leq 2T_1
\end{cases}
\end{aligned}
\right.
\label{eq:gait}
\end{equation}

In Eq. \eqref{eq:gait}, the gait time step counter $tp$ auto-increments at each fixed time step (0.01 s) and resets to 0 upon reaching $2T_1$, enabling cyclic gait periodicity. A sub-counter $tp_0$ is used for single-leg phase cycle calculation. The cosine function maps the $[0, 2\pi]$ cosine period to $[0, 2]$, which is then divided by 2 to yield a $[0, 1]$ phase activation factor, ensuring smooth action scheduling.

The factors $u_1$ and $u_2$ implement left-right leg support phase separation through an alternating activation mechanism. By combining $u_1$ and $u_2$ with joint angle coefficients (where $d_h$ denotes a dynamic adjustment coefficient and $d_0$ denotes a base angle offset), feedforward angle commands for the core lower-limb joints (hip, knee, and ankle) are generated. 
\begin{equation}
u_{\text{total}}[i] = (d_h \cdot u_f + d_0) \cdot \text{sign}(\text{idx}[i])
\label{eq:control}
\end{equation}

In Eq. \eqref{eq:control}, where $u_f$ represents $u_1$ or $u_2$, $\text{idx}$ is the index mapping array of lower limb joints, and $\text{sign}(\text{idx}[i])$ is used to control the joint movement direction. Finally, the periodic gait feedforward control of the robot is realized. Different branch skills adapt their characteristics by adjusting $T_1$, $d_h$, $d_0$, and the activation logic of $u_1$ and $u_2$.

\subsubsection{Interpolation method}
For aperiodic actions (e.g., squatting), a linear interpolation method is employed to generate reference joint angles. This method supports continuously adjustable action depth (e.g., linear mapping of squat amplitude between 0–100\%). The feedforward commands provide the RL network with baseline trajectory boundaries that carry strong physical meaning, while the policy network is responsible for optimizing action along the feedforward signal, thereby substantially improving the learning efficiency of diverse motion skills.

\subsection{Parameter Inheritance}
Parameter inheritance is primarily embodied in the training phase. The root skill (flat-ground walking) is first trained to completion, equipping the robot with stable basic locomotion capability. All branch skills are initialized from their parent skill network parameters, thus avoiding the efficiency loss associated with training from scratch.

Branch networks fully inherit the weights of the parent network as their initialization parameter value. This not only preserves the state-perception features and physical dynamics representations extracted in the hidden layers, but also avoids the random exploration inherent in training from scratch. During inference, to satisfy the memory constraints of edge devices and to fundamentally eliminate gradient interference from multi-task joint optimization, the framework employs physically isolated sub-network clusters. By sharing the same state space and similar distribution manifolds, the system ensures continuity of action commands during neural network model switching (models are in ONNX format), thereby achieving incremental expansion of multiple skills.

\subsection{Reward Shaping}

To accelerate skill convergence and improve execution stability, a set of basic reward components is first designed, as shown in Table 1.

In Table 1, $v_x$ is the robot's lateral velocity; $v_x^{\text{cmd}}$ is the robot's target lateral velocity; $v_z$ is the robot's forward velocity; $v_z^{\text{cmd}}$ is the robot's target forward velocity; $w_y$ is the robot's torso yaw angular velocity about the Y-axis; $w_y^{\text{cmd}}$ is the preset target yaw angular velocity about the Y-axis of the robot's torso; $h_{\text{target}}$ is the robot's torso target height; $\theta_{\text{roll}}$ is the body roll angle; $\theta_{\text{pitch}}$ is the body pitch angle; $u_i$ represents the output amplitude of the control command for the corresponding joint. To control redundant movements of non-essential joints, the set of penalty joints is defined as $J = \{1, 2, 5, 7, 8, 11\}$.

Since different skills have distinct requirements and preferences, the basic reward components can be recombined or supplemented with preference-specific reward terms to accommodate different skills.

\begin{table}
\caption{Basic reward components}
\label{tbl1}
\begin{tabular*}{\tblwidth}{@{}l l l@{}}
\toprule
No. & Reward Term & Mathematical Expression \\
\midrule
1 & Velocity tracking 
& $\begin{aligned}
r_{\text{vel}} &= 1 - 2|v_x - v_x^{\text{cmd}}| \\
&- 2|v_z - v_z^{\text{cmd}}|
\end{aligned}$ \\
2 & Angular velocity tracking 
& $r_{\text{wel}} = 1 - 2|w_y - w_y^{\text{cmd}}|$ \\
3 & Center of mass height 
& $r_{\text{height}} = 2(h - h_{\text{target}})$ \\
4 & Orientation constraint 
& $r_{\text{ori}} = -0.1|\theta_{\text{roll}}| - 0.1|\theta_{\text{pitch}}|$ \\
5 & Survival reward 
& $r_{\text{live}} = 1$ \\
6 & Action energy penalty 
& $r_{\text{penalty}} = -0.2 \sum_{i \in J} |u_i|$ \\
\bottomrule
\end{tabular*}
\end{table}

\subsection{Training Configuration}

\subsubsection{Network Architecture and RL Algorithm}
This study is based on the Unity ML-Agents framework, employing the Proximal Policy Optimization (PPO \cite{Schulmanetal2017}) algorithm for reinforcement learning training. For the network architecture, the actor policy network adopts a 3-layer fully connected multilayer perceptron (MLP) with 512 hidden units per layer and normalization of input state features. The policy output is set as a Gaussian distribution to ensure non-deterministic exploration. The critic value network uses a 2-layer fully connected structure (hidden layer dimension of 128) without state normalization. During training, model weights are saved every $4 \times 10^5$ steps to monitor convergence trends.

\subsubsection{Reinforcement Learning Hyperparameters}
The total training steps for different skills are adaptively adjusted according to their convergence difficulty. The core hyperparameters of the PPO algorithm employ a linear decay strategy to balance exploration in the early stages and exploitation in the later stages. The detailed configuration is presented in Table 2.

\begin{table}[htbp]
\centering
\caption{Core hyperparameter configuration of PPO}
\label{tbl2}
\begin{tabular}{@{}l l c@{}}
\toprule
No. & Parameter Name & Value \\
\midrule
1 & Time steps per update & 1000 \\
2 & Batch size & 2048 \\
3 & Replay buffer size & 20480 \\
4 & Number of epochs & 3 \\
5 & Reward scale & 1.0 \\
6 & Discount factor $\gamma$ & 0.995 \\
7 & GAE coefficient $\lambda$ & 0.95 \\
8 & Learning rate & $3 \times 10^{-4}$ \\
9 & Clipping parameter $\epsilon$ & 0.2 \\
10 & Entropy coefficient $\beta$ & $5 \times 10^{-3}$ \\
\bottomrule
\end{tabular}
\end{table}
\subsubsection{Progressive Curriculum Learning}

For complex skills with strong terrain constraints, such as stair climbing, direct training in the target staircase environment is prone to becoming trapped in local optima. To address this, an environment-variable-driven curriculum learning mechanism is introduced. Specifically, staircase step height is designated as the curriculum difficulty metric and divided into four progressive stages (Lesson 1 through Lesson 4, with heights set to 0.01 m, 0.05 m, 0.10 m, and 0.15 m, respectively). During training, when the robot’s smoothed cumulative reward at the current stage reaches a preset threshold, the environment automatically advances to the next difficulty level. This mechanism effectively alleviates the sparse reward problem on unstructured terrain, enabling smooth transition and efficient convergence of the robot’s obstacle-crossing capabilities.

\section{Comparison with Multi-Task Baseline}\label{exp}

This section selects 6 representative skills from the 11 skill types for quantitative comparison between Tree Learning and simultaneous multi-task training (Multi-task). In the experiment, the multi-task baseline uses a single policy network to train all 6 skills simultaneously, whereas Tree Learning trains each skill independently. To ensure comparable computational budgets, both experimental groups are set to a total of \(10^7\) environment steps.

The experimental results are presented in Figures \ref{fig4}–\ref{fig9}. As observed, Tree Learning achieves higher final rewards than the multi-task baseline across all 6 selected skills. Specifically, for the highly dynamic skills Jump and Run, Tree Learning attains stable rewards of 990 and 6138, respectively, while the multi-task baseline achieves only 99 and 609. Figure \ref{fig10} presents a radar chart of stable converged reward distributions across the 6 skills, clearly showing that the final reward of the Run skill under Tree Learning is substantially higher than that of the multi-task baseline.

These results indicate that simultaneous multi-task training suffers from significant gradient conflicts: the substantial differences among reward functions for different skills lead to negative transfer during optimization, rendering some highly dynamic skills (e.g., Jump) virtually unable to converge effectively. In contrast, Tree Learning fundamentally avoids gradient interference through physically isolated independent sub-networks, allowing each skill to fully leverage the neural network weights from the root skill to achieve efficient and stable convergence.

\vspace{1em}
\begin{center}
\includegraphics[width=1\linewidth]{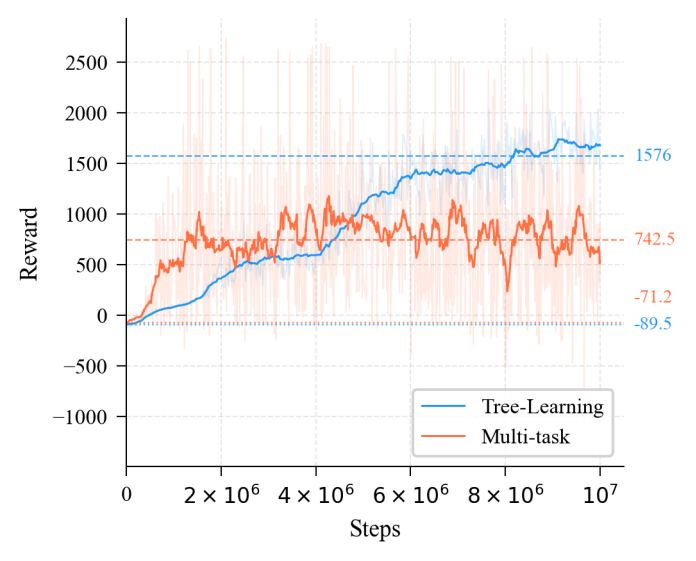}
\captionof{figure}{Reward comparison of walk skill.}
\label{fig4}
\end{center}
\vspace{1em}
\vspace{1em}
\begin{center}
\includegraphics[width=1\linewidth]{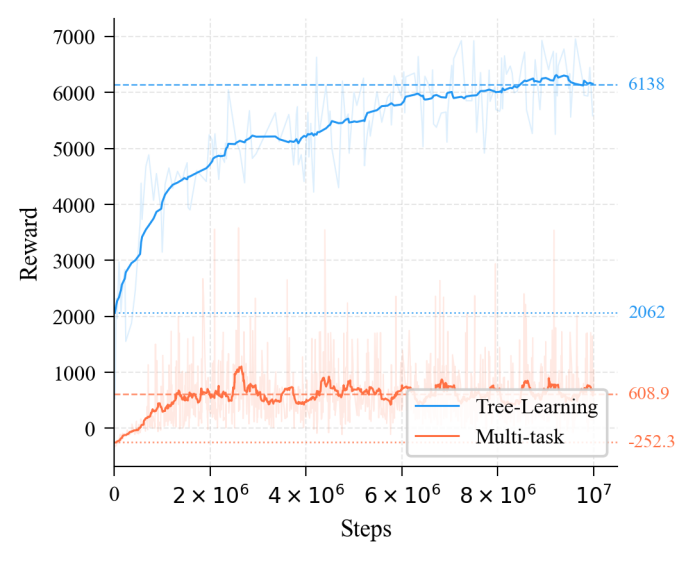}
\captionof{figure}{Reward comparison of run skill.}
\label{fig5}
\end{center}
\vspace{1em}
\vspace{1em}
\begin{center}
\includegraphics[width=1\linewidth]{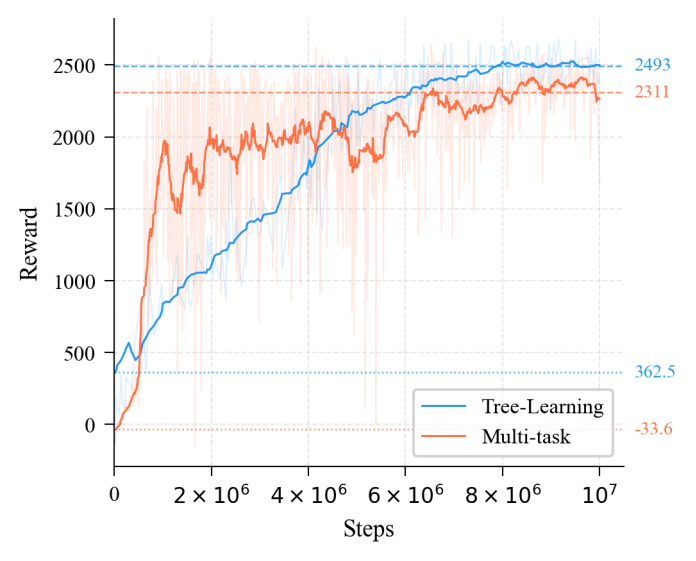}
\captionof{figure}{Reward comparison of squat skill.}
\label{fig6}
\end{center}
\vspace{1em}
\vspace{1em}
\begin{center}
\includegraphics[width=1\linewidth]{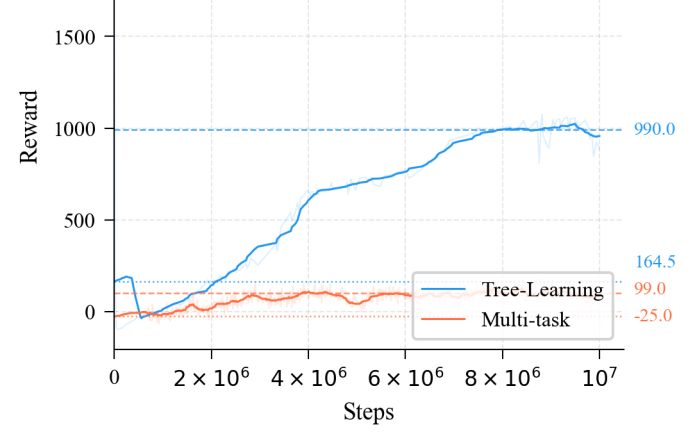}
\captionof{figure}{Reward comparison of jump skill.}
\label{fig7}
\end{center}
\vspace{1em}
\vspace{1em}
\begin{center}
\includegraphics[width=0.91\linewidth]{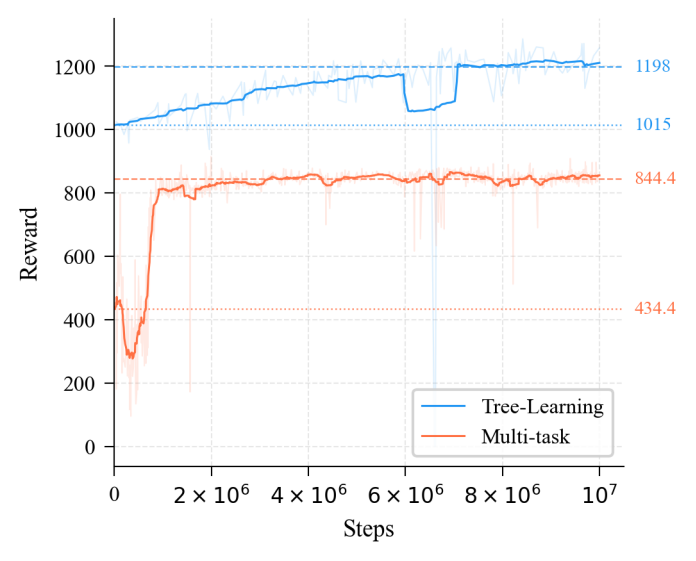}
\captionof{figure}{Reward comparison of crawl skill.}
\label{fig8}
\end{center}
\vspace{1em}
\vspace{1em}
\begin{center}
\includegraphics[width=0.91\linewidth]{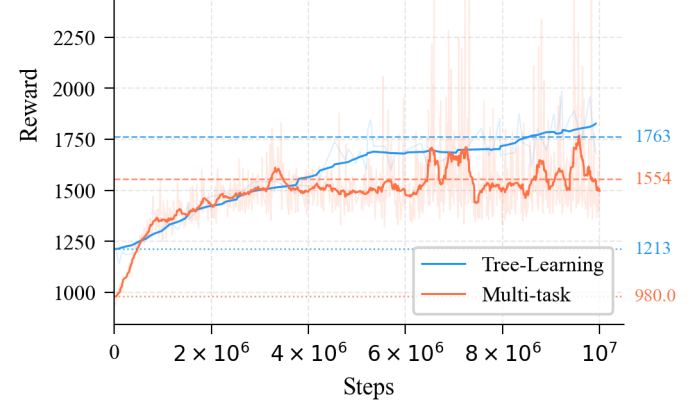}
\captionof{figure}{Reward comparison of one-leg stand skill.}
\label{fig9}
\end{center}
\vspace{1em}
\vspace{1em}
\begin{center}
\includegraphics[width=0.98\linewidth]{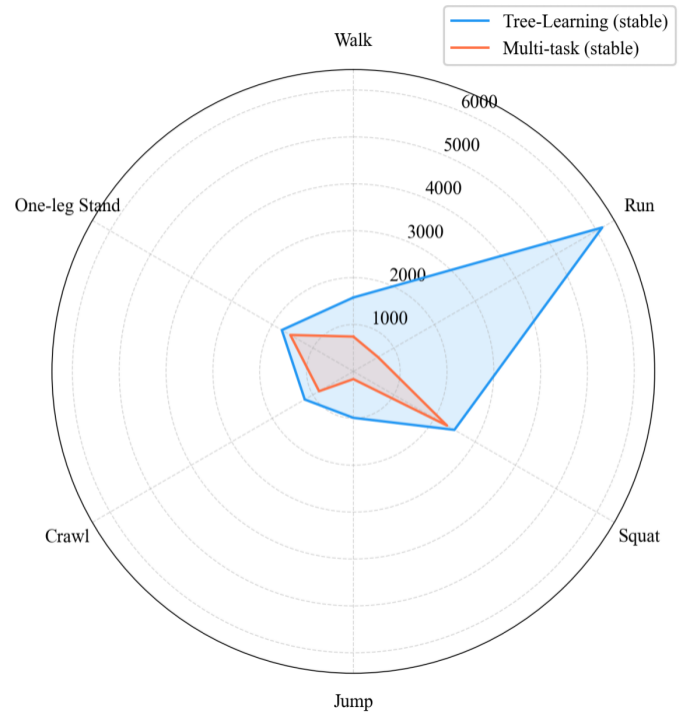}
\captionof{figure}{Final reward comparison.}
\label{fig10}
\end{center}
\vspace{1em}

\section{Super Mario Scenario Experiment}\label{exp}

\subsection{Simulation Environment and Task Setup}

\begin{figure*}[t]
\centering
\includegraphics[width=1\textwidth]{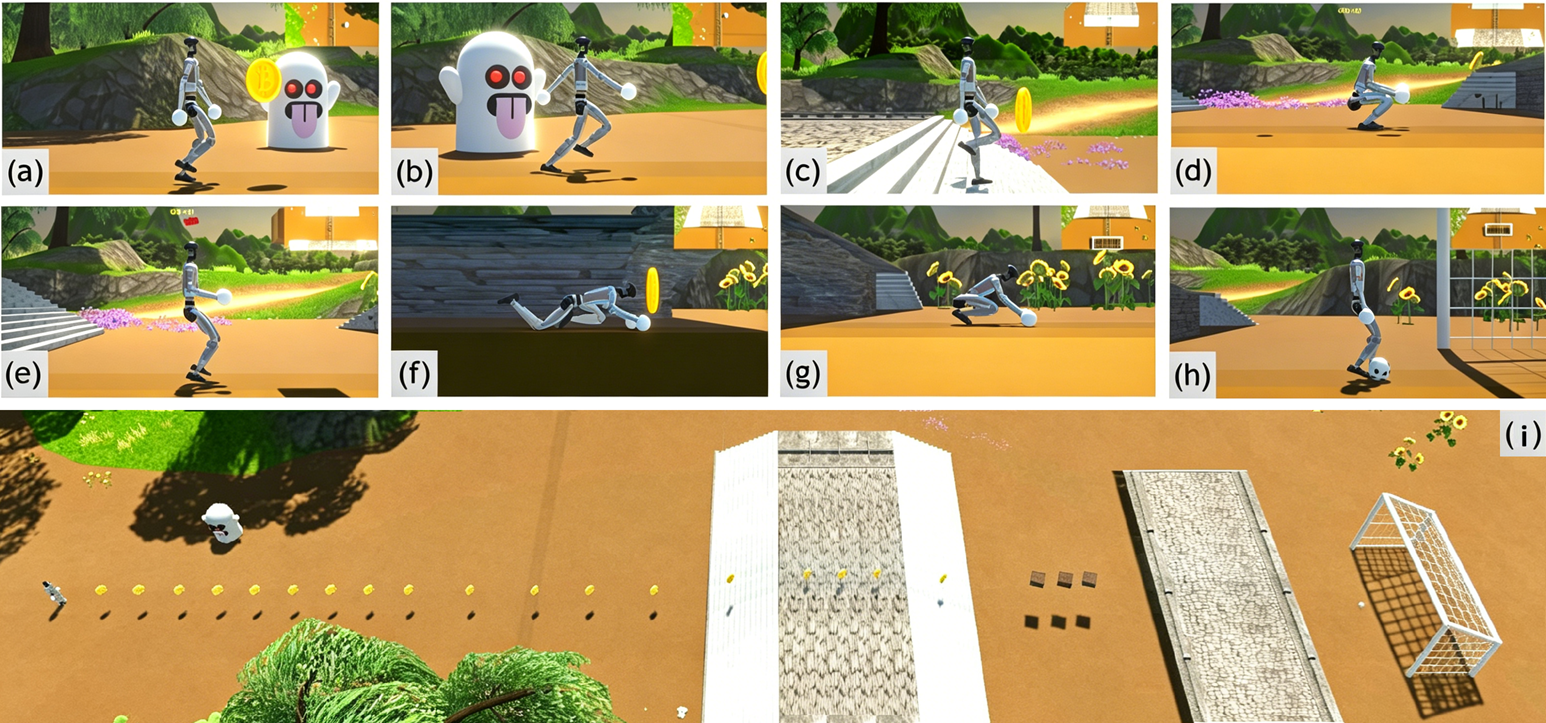}
\caption{Super Mario simulation scene. The robot sequentially performed (a) walking, (b) running to escape a ghost, (c) climbing up and down stairs, (d) lying prone, (e) crawling through a tunnel, (f) standing up, and (g) jumping to hit boxes to collect coins or gifts along the way. Finally, (h) the robot kicks a ball into the goal and wins. (i) is the global view. }
\label{fig1}
\end{figure*}

The simulation environment was built using Unity 2022.
3.62f3 LTS. The Unitree G1 humanoid robot (23-DoF version) was selected as the experimental platform, configured with 12 lower-limb joints, 1 waist joint, and 10 upper-limb joints. Multi-skill expansion was achieved through coordinated control of lower-limb joint motion and optimization of upper-limb joint motion. To increase the interestingness of the experiment and verify the comprehensive performance of multi-skill capabilities in continuous complex tasks, the experimental scenario was designed in the style of a Super Mario game, as shown in Figure \ref{fig1}. To prevent the robot from deviating laterally, two transparent walls were added to constrain movement to the forward-backward direction.

\subsection{Manual Switching Control}
The Super Mario test scene contains multiple interactive constraints including pursuers, coins, and complex terrain, posing high demands on the temporal sequencing of the humanoid robot's actions. Experimental observations revealed that the robot must complete multiple action transitions within extremely narrow action windows. Through manual switching control via keyboard, a typical full-process task sequence was successfully executed, validating the physical feasibility of the Tree Learning framework in handling highly dynamic and tightly coupled tasks. See Supplementary Video 1 for details.

\subsection{Automatic Switching Control}
For convenience, we also developed a method to enable the robot to automatically switch from walking to running and subsequent complex action sequences based on environmental cues. To quantitatively assess switching continuity, center-of-mass (CoM) height and forward velocity curves were extracted during the automatic switching process.

\vspace{1em}
\begin{center}
\includegraphics[width=1\linewidth]{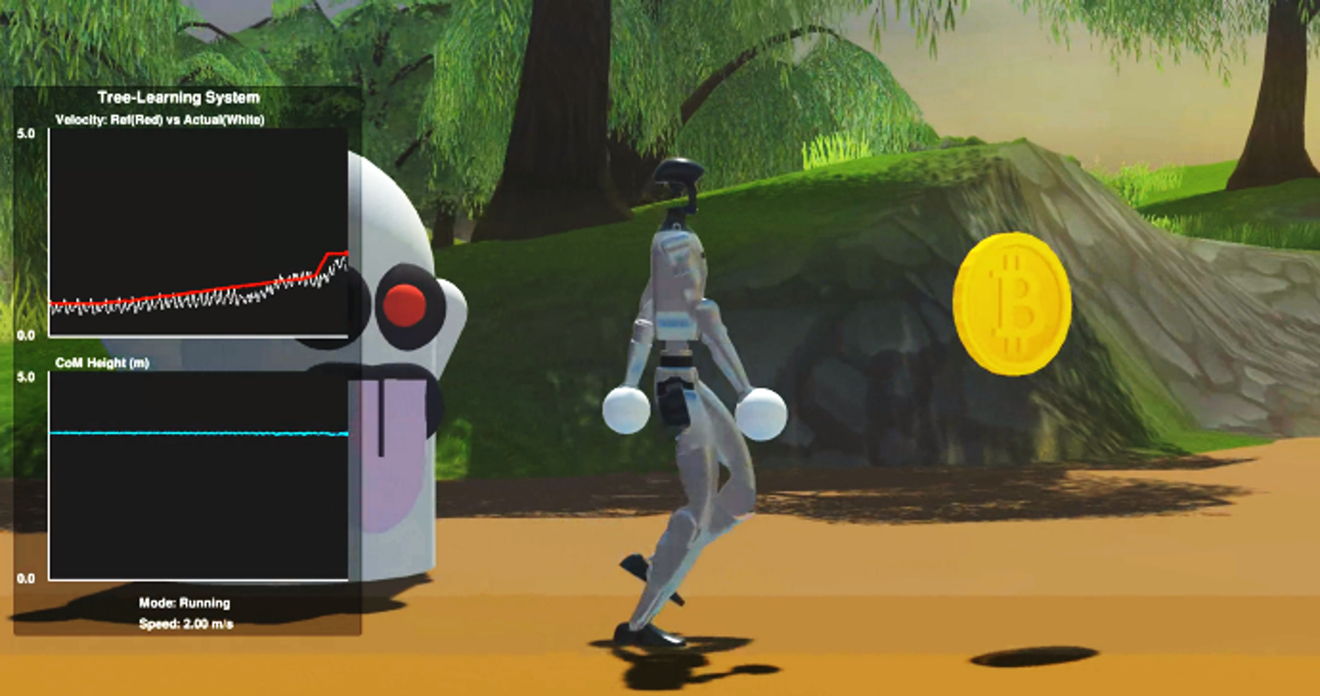}
\captionof{figure}{Ghost chasing scene.}
\label{fig11}
\end{center}
\vspace{1em}

As shown in Figure \ref{fig11}, when the robot detects a ghost pursuit and triggers an acceleration command, the CoM velocity curve exhibits notable continuity, smoothly transitioning from 0.8 m/s to 2.0 m/s without any step-like jumps or acceleration spikes throughout the switching interval. This smooth locomotion performance physically validates the effectiveness of the state space consistency design proposed in Section 2—namely, that physical quantities from different skill models can naturally align at the moment of switching. The entire automatic task flow demonstrates that the framework possesses high kinematic coherence when handling heterogeneous skill sequences. See Supplementary Video 2 for more details.

\section{Autonomous Navigation Experiment}\label{exp}

To further validate the robustness and control quality of the Tree Learning framework under a prolonged continuous task, an autonomous navigation task was designed in a classical Chinese garden scenario. Unlike the manual and automatic switching experiments in previous section, which focused on verifying switching feasibility, this experiment concentrates on a comprehensive evaluation of the robot during nearly 240 seconds of continuous locomotion, encompassing path planning, velocity control, gait switching, posture balancing, and forward-turn exchange. The results are given in Supplementary Video 3. The sampling rate was set to 50 Hz, resulting in about 12,000 recorded frames.

The navigation test scenario includes buildings, streets, and bridges, as shown in Figure \ref{fig12}. Based on the skills learned by Tree Learning, we adopted a navigation workflow as shown in Figure \ref{fignav}. Figure \ref{fig12} shows the top-view trajectory of the robot. Starting from the starting point (green dot), the robot follows the path from the planner to sequentially approach multiple operator-specified target points and finally reach the end point (red square). The trajectory covers an area of approximately 30 m × 30 m. Multiple turns and obstacle avoidance behaviors are observed along the path, indicating that the navigation system can effectively plan feasible paths in complex scenarios with various terrains.

During the experiment, the operator clicks random target points in the scenario in real time. After the robot reaches the current target, the next target point is clicked immediately. During the task, the robot sequentially experiences three motion modes within 240 seconds: walking (Walk), running (Run), and stair climbing (Stair). The gait mode switching is automatically performed by the navigation system according to environmental conditions: when the target distance is far, the system automatically switches to Run mode to improve locomotion efficiency; when stairs are detected ahead, the system automatically switches to Stair mode; in other cases, Walk mode is used by default.

\vspace{1em}
\begin{center}
\includegraphics[width=1\linewidth]{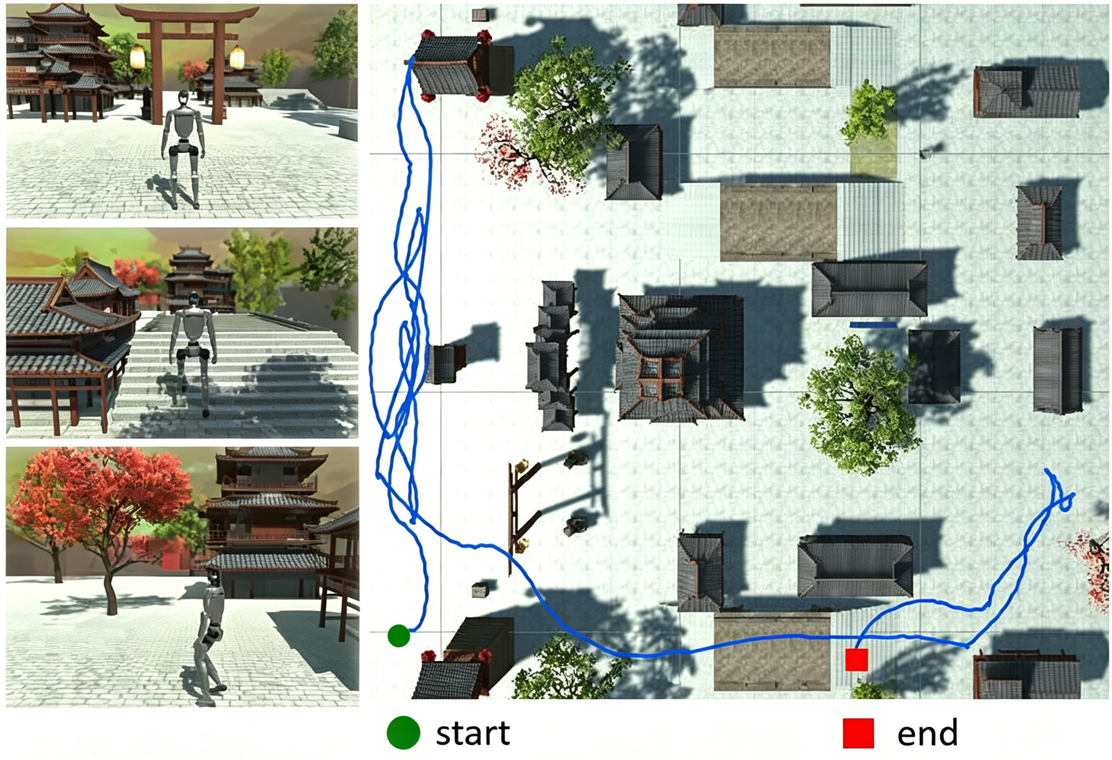}
\captionof{figure}{Autonomous navigation task. The left shows snapshots during the task. The right shows the top-down trajectory of the robot.}
\label{fig12}
\end{center}
\vspace{1em}

\vspace{1em}
\begin{center}
\includegraphics[width=0.98\linewidth]{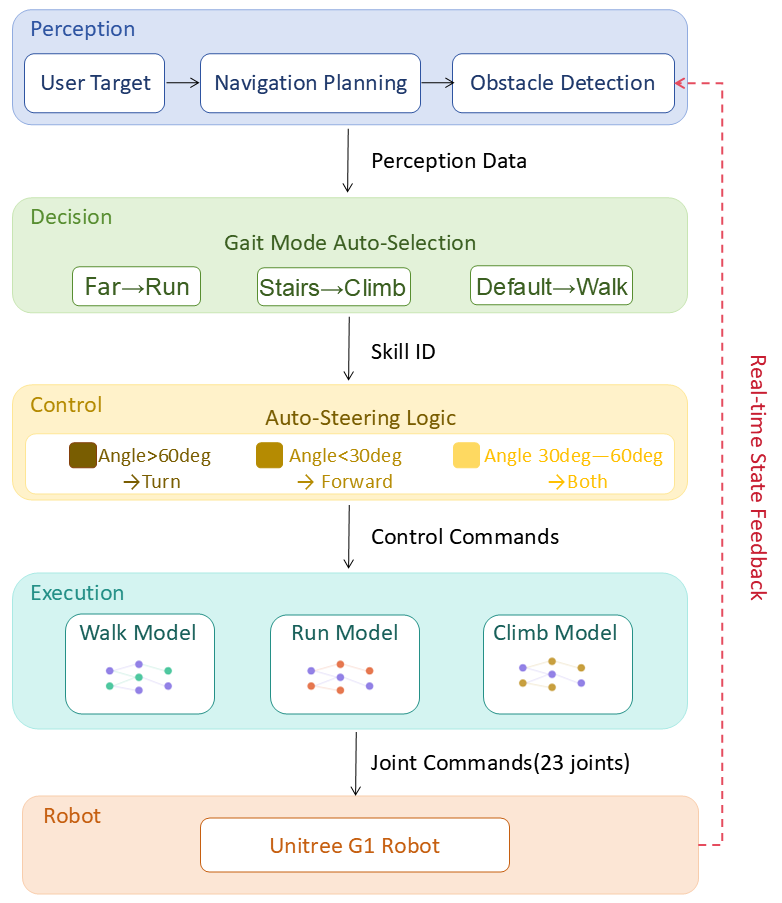}
\captionof{figure}{Autonomous navigation workflow.}
\label{fignav}
\end{center}
\vspace{1em}

\subsection{ Navigation Performance Analysis}

Figure \ref{fig13} presents the position curves of the robot's center of mass on the three coordinate axes over time. The X-axis position continuously rises from 0 to approximately 27 m after 120 s, corresponding to the robot's long-distance navigation phase toward a far target. The Z-axis position exhibits reciprocating fluctuations, reflecting the robot's path between multiple target points. Of particular interest is the Y-axis (height) data: during the flat-ground phase from 0 to 160 s, the center-of-mass height remains stable at approximately 0.78 m with minimal fluctuations. In the interval from 170 to 190 s, the center-of-mass height rises sharply from 0.78 m to approximately 1.78 m and then stabilizes at the top of the stairs, proving that the robot successfully climbed a stair of about 1 m in height. The time interval of this height jump matches exactly with the Stair mode activation interval in Figure \ref{fig14}, verifying the physical effectiveness of gait switching from the spatial position dimension. At the end of the experiment (around 235 s), the Y-axis shows a slight rise again, corresponding to the second short stair climbing.

Figure \ref{fig14} quantitatively presents the navigation performance metric over time. The upper figure shows the target distance curve, which exhibits a typical ``sawtooth'' descending pattern: whenever the distance approaches zero, the system determines that the current target has been reached, the operator specifies a new target immediately, and the distance jumps accordingly. The green vertical lines mark the events of reaching each target. During the entire experiment, the robot successfully reached 22 target points with only one stuck recovery, demonstrating extremely high navigation reliability. Notably, a long-distance navigation segment (with a distance peak close to 20 m) appears around 120 s, which is a typical scenario where the system automatically switches to Run mode based on the target distance. The lower figure shows the obstacle factor curve, which reflects the degree to which obstacles in front suppress the robot's speed: a value of 1.0 indicates no obstacles ahead, while values close to 0 indicate detection of close-range obstacles. In the early half of the experiment (0–120 s), the obstacle factor remains at 1.0, indicating that this area is dominated by flat ground or open corridors. In the intervals of 140–170 s and 200–240 s, the obstacle factor drops multiple times, corresponding to the robot entering stairwells or narrow passages and performing obstacle avoidance deceleration.

Figure \ref{fig15} presents the statistic of the navigation experiment in three forms: pie chart, bar chart, and histogram. The gait distribution pie chart shows that Walk mode accounts for 87.8\%, Stair mode for 10.1\%, and Run mode for 2.0\%, which is consistent with the expected design of mainly normal walking, automatic acceleration for long-distance targets, and stair climbing. The navigation bar chart visually compares the completion of 22 target points with only one stuck recovery, verifying the high reliability of the system in multi-terrain environments. The forward velocity histogram shows a bimodal distribution, with the main peak near 0.0 m/s (corresponding to standing or deceleration phases) and the secondary peak in the range of 0.8–1.0 m/s (corresponding to steady-state forward movement). This distribution is highly consistent with the designed auto-steering control logic.

\vspace{1em}
\begin{center}
\includegraphics[width=1\linewidth]{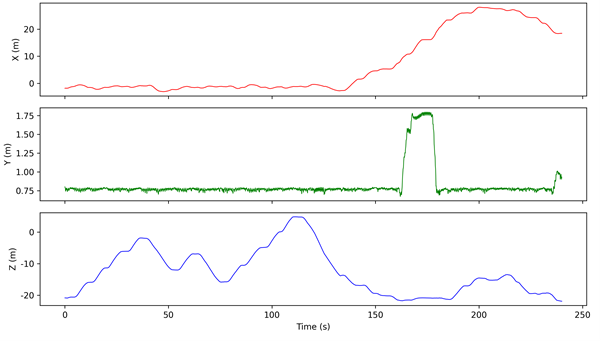}
\captionof{figure}{Robot CoM position curve.}
\label{fig13}
\end{center}
\vspace{1em}

\vspace{1em}
\begin{center}
\includegraphics[width=1\linewidth]{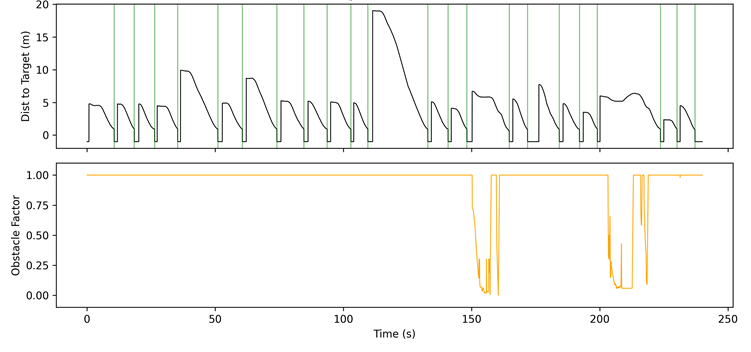}
\captionof{figure}{Autonomous navigation metric.}
\label{fig14}
\end{center}
\vspace{1em}

\vspace{1em}
\begin{center}
\includegraphics[width=1\linewidth]{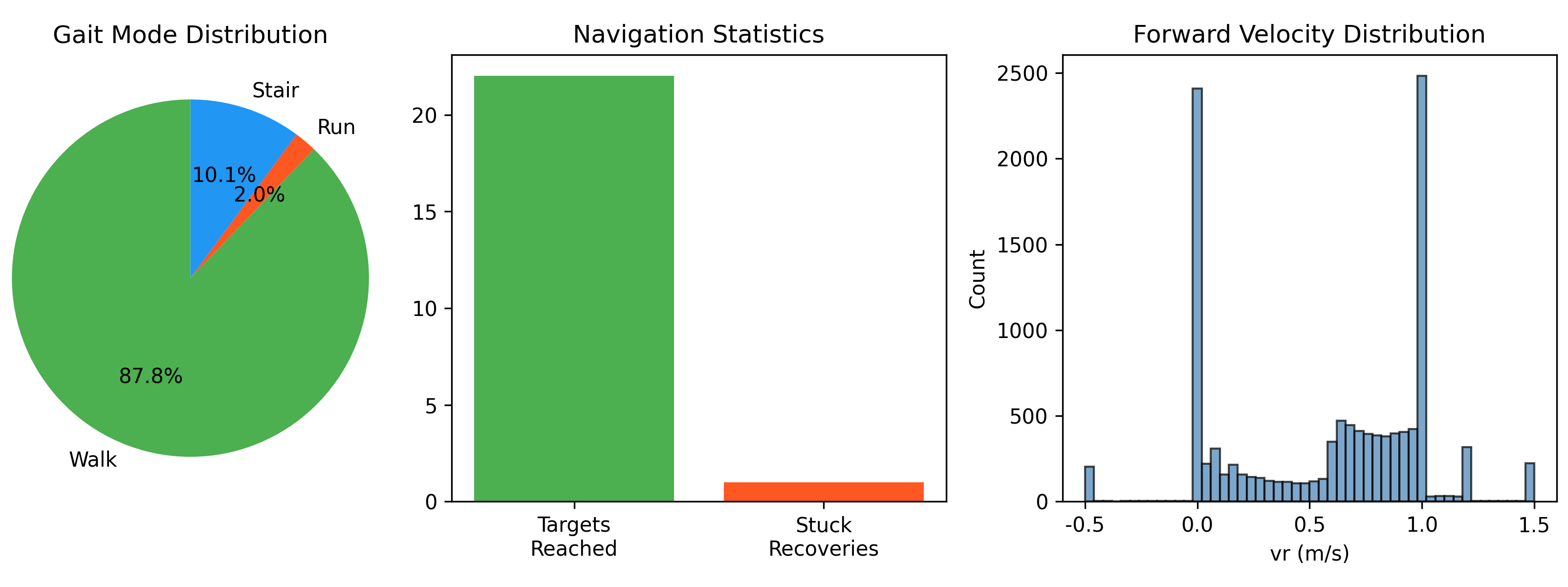}
\captionof{figure}{Autonomous navigation statistic.}
\label{fig15}
\end{center}
\vspace{1em}

\subsection{Motion Control Analysis}

% Figure \ref{fig16} shows the actual velocity of the robot's center of mass in the world coordinate system. $v_x$ and $v_z$ correspond to the velocities in two orthogonal directions on the horizontal plane. Their amplitudes increase significantly after 15 s, becoming more pronounced when the robot enters the long-distance navigation and stair regions. The vertical velocity component remains within ±0.3 m/s during the flat-ground phase, reflecting the periodic undulation of the center of mass during walking. In the stair region (170–190 s), the vertical velocity shows a negative peak rebound, corresponding to the vertical impact when the robot descends the stairs. By comparing the actual center-of-mass velocity with the command velocity in Figure \ref{fig17}, it can be found that although the actual velocity contains high-frequency jitter due to physical interactions, its low-frequency trend is consistent with the command velocity sequence, indicating that the skill models under the Tree Learning framework have good velocity tracking capability for control commands.

% \vspace{1em}
% \begin{center}
% \includegraphics[width=1\linewidth]{figs/16.png}
% \captionof{figure}{Robot CoM Velocity.}
% \label{fig16}
% \end{center}
% \vspace{1em}

Figure \ref{fig17} shows the time series of control commands $v_r$ (forward velocity), $w_r$ (turning angular velocity). The $v_r$ curve exhibits a distinct "step-pulse" alternating pattern: $v_r$ stabilizes at 0.8–1.0 m/s during straight segments and drops rapidly to near zero during turning segments. Correspondingly, the $w_r$ curve fluctuates significantly during turning segments (peak value ±1.0 rad/s) and remains small during straight segments. Notably, $v_r$ shows a peak of 1.5 m/s at around 120 s, corresponding to the high-speed forward phase when Run mode is activated. The inverse coupling relationship between $v_r$ and $w_r$ visually verifies the primary auto-steering control logic of forward and turning in the navigation system: the forward velocity is preferentially reduced during turning to ensure steering accuracy, while the forward efficiency is maximized during straight-line movement.

\vspace{1em}
\begin{center}
\includegraphics[width=1\linewidth]{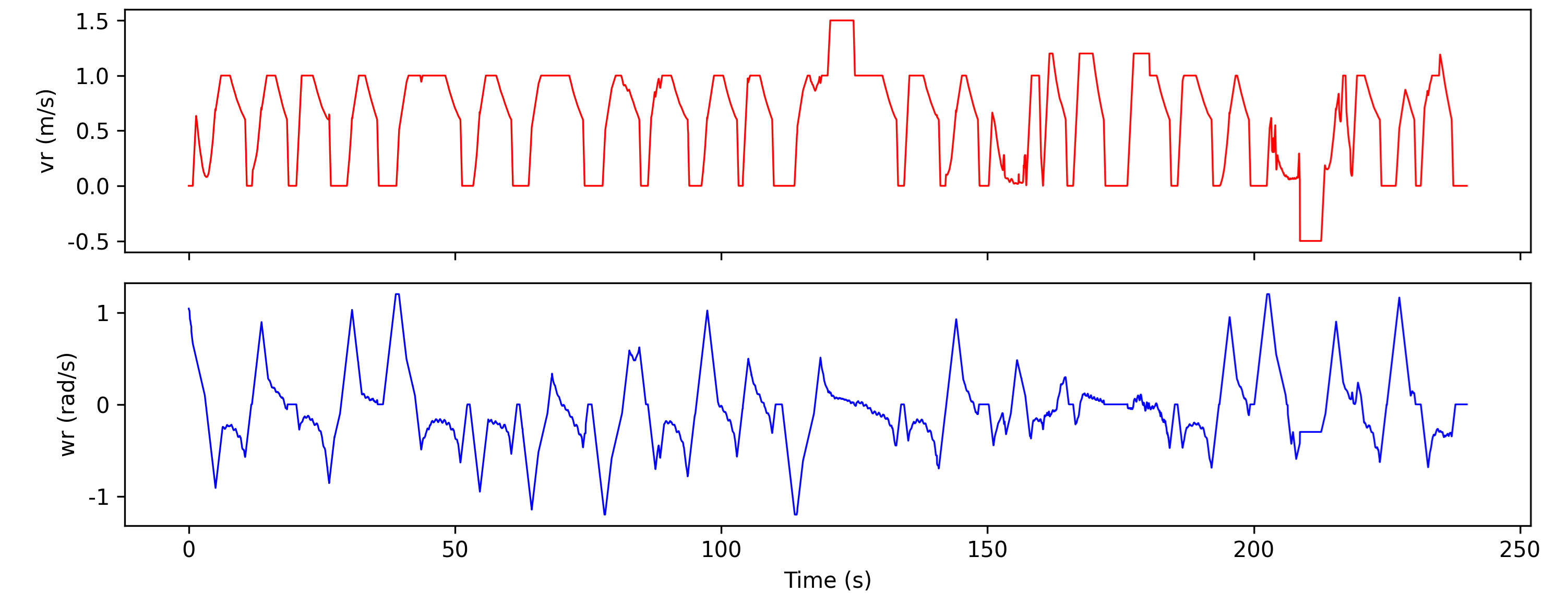}
\captionof{figure}{Control commands(vr/wr).}
\label{fig17}
\end{center}
\vspace{1em}

To further reveal the inherent structure of the forward-turn mutual exclusion relationship, Figure \ref{fig18} plots the $v_r$-$w_r$ phase diagram with target angle as the color mapping. The gray dashed line in the figure represents the speed-turn interlock boundary, which is determined by the parameters in the controller. Two distinct regions can be clearly observed in the figure: when the target angle is large (warm color, angle > 60°), the data points are concentrated in the region of low $v_r$ and high $|w_r|$ (marked as ``Turn-dominant'' in the figure), indicating that the robot prioritizes turning to align with the target direction; when the target angle is small (cool color, angle < 30°), the data points are concentrated in the high $v_r$ and low $|w_r|$ region (marked as ``Forward-dominant'' in the figure), indicating that the robot has basically aligned with the target and accelerates forward. The entire data distribution is enclosed by the interlock boundary, and the control commands are always within the physical safety envelope regardless of the gait mode.

\vspace{1em}
\begin{center}
\includegraphics[width=1\linewidth]{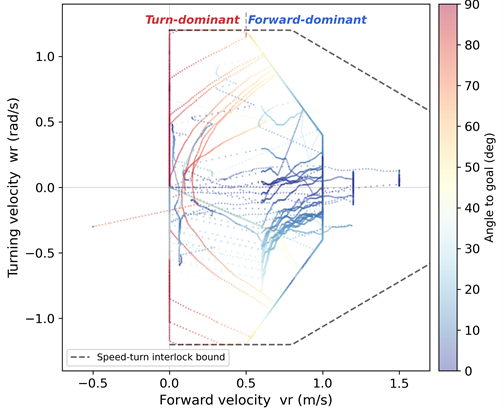}
\captionof{figure}{vr-wr phase portrait.}
\label{fig18}
\end{center}
\vspace{1em}

\subsection{Gait Switching and Posture Stability}

Figure \ref{fig21} shows the temporal switching of gait modes during the 240-second experiment in the form of a color band diagram. The robot remained in Walk mode throughout the 0–120 s period, conducting stable locomotion in the flat environment. Then, at approximately 120 s, as the operator specified a farther target point, the system automatically switched to Run mode (lasting about 5 s) for acceleration, and then returned to Walk mode. In the 160–180 s interval, the robot reached the stair area, and the system automatically switched to Stair mode to adapt to the step terrain, returning to Walk mode after passing the stairs. At the end of the experiment (around 235 s), the robot encountered stairs again and briefly entered Stair mode. The gait switching sequence throughout the experiment was highly consistent with terrain changes and target distances, and all switching was completed automatically by the navigation system, indicating that the multi-skill switching logic under the Tree Learning framework has good environmental adaptability in autonomous navigation scenarios.

Figure \ref{fig22} presents the posture stability data of the robot's torso. The upper figure shows the time curves of roll and pitch angles. During the flat-ground walking phase, both roll and pitch angles were controlled within about ±2.5°, demonstrating high posture stability. In the stair interval (160–180 s), the Roll angle fluctuations increased to approximately ±(5–7)°, caused by periodic disturbances from the stair steps, but remained within a safe range without falls or severe instability. This further verifies the design concept of a unified state space shared by all skill models in the Tree Learning framework: different branch networks can achieve continuous transition of action commands at the moment of switching, ensuring posture robustness throughout autonomous navigation.

\vspace{1em}
\begin{center}
\includegraphics[width=1\linewidth]{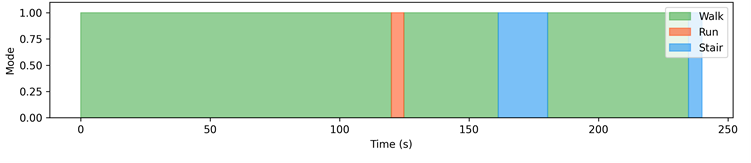}
\captionof{figure}{Gait mode distribution along path.}
\label{fig21}
\end{center}
\vspace{1em}

\vspace{1em}
\begin{center}
\includegraphics[width=1\linewidth]{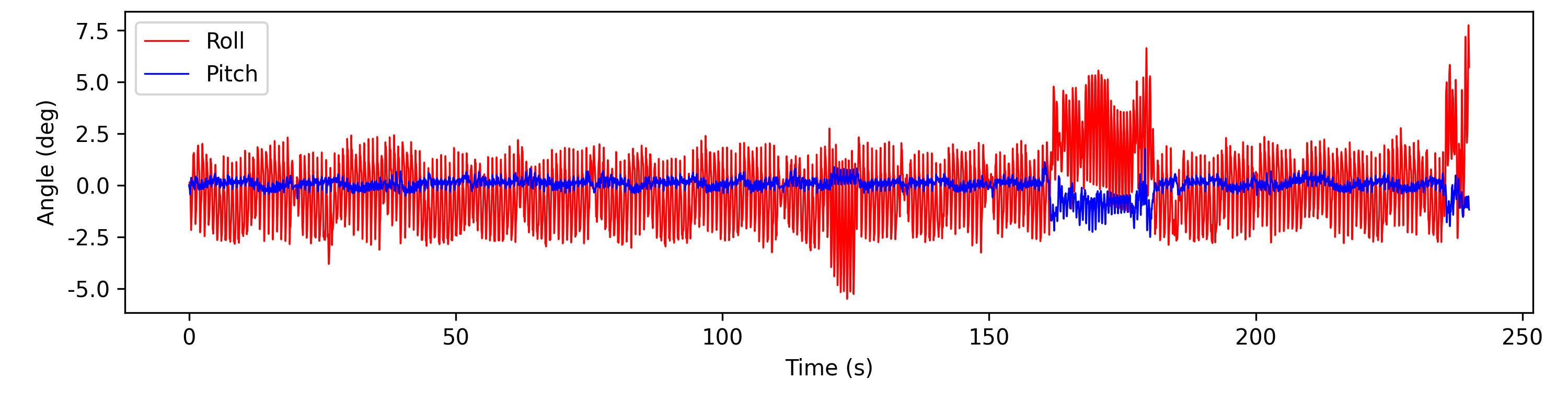}
\captionof{figure}{Torso orientation curve.}
\label{fig22}
\end{center}
\vspace{1em}

% \subsubsection{Subsection Summary}

% Through the 240-second autonomous navigation experiment conducted in a classical Chinese garden scenario, the Tree Learning framework was comprehensively validated across five dimensions: trajectory planning, navigation efficiency, motion control, gait switching, and posture stability. The experimental results demonstrate that: (1) the robot successfully reached 22 target points with only 1 stuck recovery, exhibiting extremely high navigation reliability; (2) forward velocity vr and turning angular velocity wr exhibited a clear mutual exclusion control pattern, with all control commands constrained within the speed-turn interlock boundary; (3) the Walk, Run, and Stair gait modes switched automatically based on target distance and terrain features, forming differentiated distributions in the vr-wr phase space, confirming that each branch skill learned independent and reasonable control strategies; (4) torso Roll and Pitch angles remained within safe ranges throughout (flat ground ±2°, stairs ±7°), with no posture transients at gait switching moments, validating the posture robustness of the Tree Learning framework for seamless multi-skill switching.
\section{Discussion}\label{discussion}

\subsection{Mechanism Effectiveness Analysis}

The Tree Learning framework achieves both enhanced multi-skill expansion efficiency and high motion stability, primarily owing to the synergistic effect of its underlying design components. First, the parameter inheritance mechanism reuses the neural network weights embedded in the root skill (e.g., flat-ground walking). This approach of learning by building upon established foundations substantially reduces the blind exploration range in the high-dimensional state-action space, significantly lowering the exploration cost and sample complexity for new skills. Second, the feedforward action and reward shaping strategy provides targeted dense guidance for different branch skills, not only accelerating network convergence but also ensuring action reasonableness under physical constraints. Third, the multi-modal action adaptation mechanism combining phase modulation and trajectory interpolation enables the framework to accommodate both periodic gaits and aperiodic large-range motions, substantially extending the system’s applicability boundary.

\subsection{Summary of Core Advantages}
Compared with conventional reinforcement learning-based motion control approaches, the Tree Learning framework proposed in this paper demonstrates advantages as follows:

(1) High training efficiency: Owing to hierarchical parameter reuse, the converged rewards of Tree Learning consistently surpass those of the multi-task baseline when expanding new skills.

(2) Avoidance of catastrophic forgetting: Through the physically isolated tree-structured network storage topology, the system achieves decoupling of old and new skills, with a retention rate of 100\% for existing fundamental skills, resolving the negative transfer problem in single-network multi-task learning.

(3) Broad scenario adaptability: The framework successfully covers a wide skill spectrum ranging from simple flat-ground locomotion to complex environments (e.g., stairs, tunnel) and highly dynamic interactive motion (e.g., ball kicking, jumping).

(4) Practical interactivity: Through a lightweight global clock and state relay logic, the system supports smooth, ultra-low-latency dynamic skill switching under user commands, with no posture transients or fall instabilities observed.

\subsection{Limitations and Future Directions}
Despite the promising results achieved in simulation, this study has certain limitations that point to directions for future work. On one hand, the current framework has been validated only in the Unity simulation environment, without fully accounting for the complexities of real motor dynamics, sensor noise, and contact friction—constituting an objective Sim-to-Real gap. In future work, we plan to add domain randomization and implicit state estimation techniques to pursue zero-shot transfer deployment to physical humanoid robots. On the other hand, long-period complex motions (e.g., dance) have not yet been addressed. The next step will involve developing automated motion retargeting and keyframe auto-extraction algorithms based on human video demonstrations, to further improve the construction efficiency of complex skills.

\section{Conclusion}\label{conclusion}
This paper addresses the prevalent pain points of low training efficiency and susceptibility to catastrophic forgetting in multi-skill learning for humanoid robots by proposing Tree Learning, a multi-skill continual learning framework. The framework innovatively adopts a root-branch hierarchical network architecture, achieving efficient expansion of 11 diverse motor skills for humanoid robots through prior parameter inheritance, task-level reward shaping, and multi-modal action feedforward adaptation. Simulation experiments demonstrate that, compared with the simultaneous multi-task training baseline, Tree Learning not only improves the converged rewards for new skills but also achieves seamless human-robot interactive switching of cross-terrain, cross-modal action commands. This work fundamentally avoids the gradient conflicts arising from complex skill accumulation from a mechanistic perspective, providing an efficient system-level solution for continual learning of embodied intelligent agents.

\section*{Acknowledgements}
This work was supported by the Science and Technology Commission of Shanghai Municipality (24511103304).

\printcredits

\section*{Declaration of competing interest}
The authors declare that they have no known competing financial interests or personal relationships that could have appeared to influence the work reported in this paper.

\bibliographystyle{cas-model2-names}

\bibliography{cas-refs}


\end{document}